\documentclass[letterpaper, 10 pt, conference]{ieeeconf}  

\IEEEoverridecommandlockouts                              

\usepackage[sorting=none]{biblatex}  
\usepackage{graphics}
\usepackage{amsfonts}
\usepackage{amsmath}

\usepackage{amssymb}
\usepackage[colorlinks, linkcolor=red, citecolor=blue]{hyperref}
\usepackage{algorithm}
\usepackage{algpseudocode}
\usepackage{epsfig}
\usepackage{graphicx}
\usepackage{caption}
\usepackage{subfigure}
\usepackage{mathrsfs}
\usepackage{array}
\usepackage{multirow}
\usepackage{multicol}
\usepackage{booktabs}
\usepackage{makecell}
\usepackage{xcolor}
\usepackage{amsfonts}
\usepackage{gensymb}
\usepackage{flushend}

\title{\LARGE \bf
Neural Motion Prediction for In-flight Uneven Object Catching
}

\author{Hongxiang Yu, Dashun Guo, Huan Yin, Anzhe Chen, Kechun Xu, Yue Wang and Rong Xiong
\thanks{$^{1}$All authors are with the State Key Laboratory of Industrial Control and Technology, and the Institute of Cyber-Systems and Control, Zhejiang University, Hangzhou 310058, China.}%
}

\begin{document}

\maketitle
\thispagestyle{empty}
\pagestyle{empty}

\begin{abstract}

In-flight objects capture is extremely challenging. The robot is required to complete trajectory prediction, interception position calculation and motion planning in sequence within tens of milliseconds. As in-flight uneven objects are affected by various kinds of forces, motion prediction is difficult for a time-varying acceleration. In order to compensate the system's non-linearity, we introduce the Neural Acceleration Estimator (NAE) that estimates the varying acceleration by observing a small fragment of previous deflected trajectory. Moreover, end-to-end training with Differantiable Filter (NAE-DF) gives a supervision for measurement uncertainty and further improves the prediction accuracy. Experimental results show that motion prediction with NAE and NAE-DF is superior to other methods and has a good generalization performance on unseen objects. We test our methods on a robot, performing velocity control in real world and respectively achieve 83.3$\%$ and 86.7$\%$ success rate on a ploy urethane banana and a gourd. We also release an object in-flight dataset containing 1,500 trajectorys for uneven objects.

\end{abstract}

\section{INTRODUCTION}

In-flight objects capture is extremely challenging especially for uneven objects. In tens of milliseconds, the robot needs to successively predict the flight trajectory, calculate a feasible interception position and plan the arm’s motion\cite{kim2014catching,chen2017hitting,salehian2016dynamical}. Among these tasks, the motion prediction is quite critical for the whole catching system. With the predicted trajectory via motion prediction, the arm can be set to the interception position in a suitable catching configuration before the arrival of objects. Otherwise, imprecise prediction with accumulated errors will lead to a imperfect interception position, thus resulting in low success rate of objects capture.\par



\begin{figure}[htbp]
\centering
\includegraphics[scale=0.27]{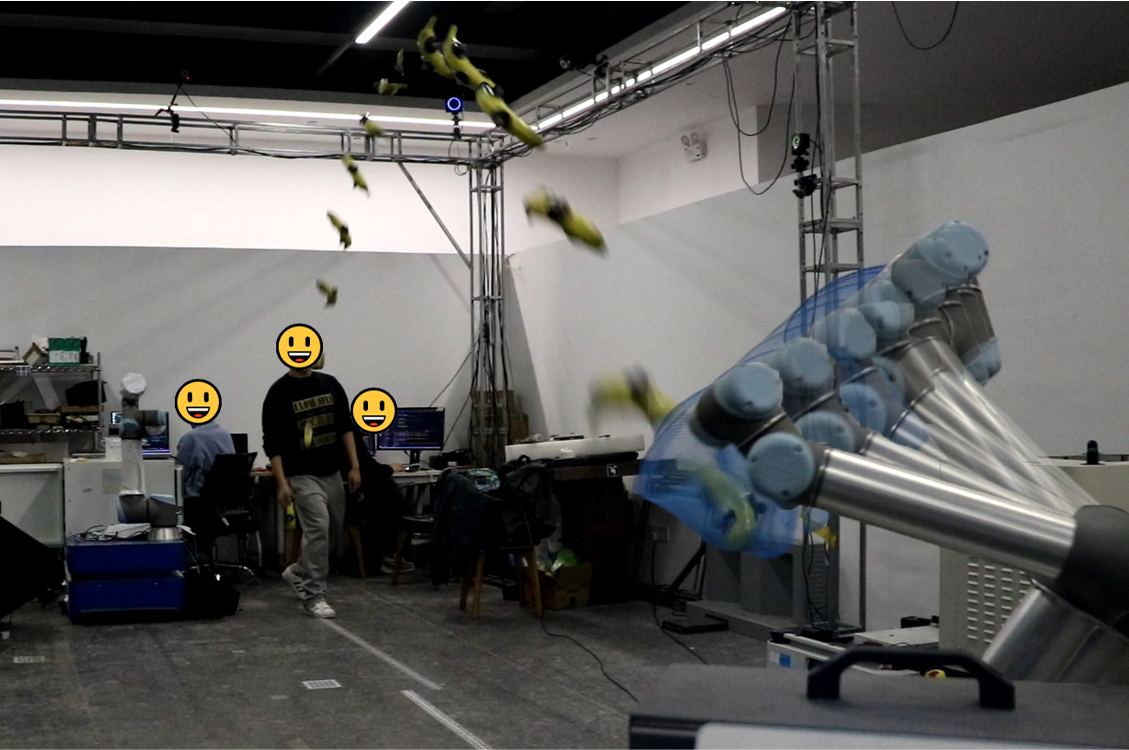}
\caption{An uneven banana is thrown towards an UR5 robot 5 meters away at a velocity of 5-6$m/s$, and the banana only flies for 1$s$ before fallen into UR5's workspace. By predicting banana's flying trajectory with NAE-DF, the UR5 catches the banana successfully. }
\label{fig:teaser}
\end{figure}

Most of robotic systems are able to catch an even object such as a ball \cite{dong2020catch,bauml2011catching,bauml2010kinematically,lampariello2011trajectory,zhang2014spin}, while few studies focus on motion prediction of uneven objects with complex aerodynamics. In these existing studies, they either regard the trajectory of the ball as a parabola \cite{hong1997experiments,riley2002robot}, or simplify the ball as a solid particle subjects to aerodynamic force and gravity \cite{chen2017hitting,frese2001off,muller2011quadrocopter}. For example, by formulating the differential equation of the ball's motion, some research works \cite{barker1995bayesian} use the vision system to track the solid ball and then build Extended Kalman filter for state correction. However, we can not transfer the modelling of the ball to uneven object directly, since the dynamics of uneven object is more complicated. \par

Current modeling for uneven objects can be divided into two categories, physics-based mechanism modeling and traditional machine learning methods. Physics-based modeling\cite{jia2019batting,gardnermotion} requires prior information such as mass, position of COM, moment of inertia, and shape of the object, etc. For uneven objects, which have irregular shapes and intricacy aerodynamics, it is troublesome to collect such prior information. When dealing with new objects, these prior information needs to be re-measured or even re-modeled again. As for learning based methods, several works \cite{kim2014catching,kim2012estimating,salehian2016dynamical} propose to use bundle of machine learning methods to estimate non-linear dynamics after observing some samples of trajectories, which achieve better performance compared to the physics-based modeling. However, the number of learning parameters in kernel methods is highly related to the volume of training data, it cannot be trained with large scale data, thus leading to the limited performance when generalized to unseen objects. 


\begin{figure*}[t]
\centering
\includegraphics[width=16cm]{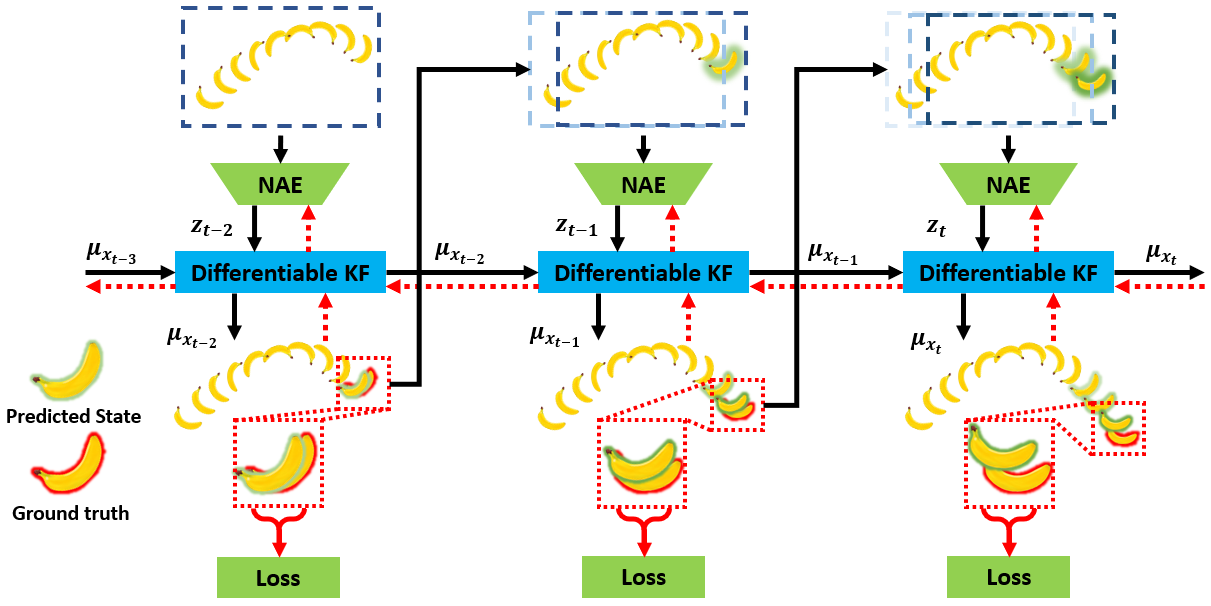}
\caption{A integral illustration of NAE-DF. NAE-DF is a Differentiable Filter embedded with a Neural Acceleration Estimator as the  measurement model. The entire framework is trained end-to-end, where the black lines indicating the flowing of data and red dash line indicating the backpropagation of loss.}
\label{fig:NAE-DF}
\end{figure*}

According to the above analysis, we consider that traditional machine learning methods should implicitly capture information associated with external force to show good performance. But they need large model, as they have to capture both temporal and spatial variations, resulting in high complexity. Therefore, we borrow part of the physical idea to formulate the object as a dynamic system for temporal modeling. Then we apply learning method to explicitly estimate acceleration at only one timestep as the system spatial measurement, which is a manifestation of force. Specifically, we propose a neural estimator utilizing a small fragment of previous deflected trajectory to predict the acceleration, named \textbf{Neural Acceleration Estimator (NAE)}. Finally, we propose a Differentiable Kalman Filter to filter the state by optimally fusing NAE measurements and temporal dynamics, named \textbf{NAE-DF}. Thanks to the differentiability, the whole architecture can be trained in an end-to-end manner\cite{haarnoja2016backprop,yin2020rall,lee2020multimodal}. In this way, the uncertainty of the NAE measurement can also be indirectly supervised. Such modeling injects the inductive bias to the learning architecture, thereby can be expected to improve the generalization performance.\par

Experimental results show that the proposed NAE and NAE-DF are superior to existing methods in prediction accuracy and generalization performance for uneven object in both public dataset and real world. Fig.\ref{fig:teaser} shows a demonstration of our catching experiment in real world, and we achieve a success rate of 83.3$\%$ for a ploy urethane banana and 86.7$\%$ when generalizing the trained banana's model to an unseen gourd.\par

Overall, the contributions of this paper are three-fold:
\begin{itemize}
\item We propose a Neural Acceleration Estimator which makes accurate estimation to the time-varying acceleration caused by various kinds of external forces, without any prior information like mass, shape or inertia. Then we can estimate the trajectories of in-flight uneven objects with a linear model.
\item We embed NAE into a Differentiable Filter and train the NAE-DF in an end-to-end manner, thus supervising the uncertainty of measurement models. Compared to NAE, NAE-DF achieves a better performance for motion prediction on uneven objects.
\item The third contribution is the real world validation of NAE and NAE-DF. We use a UR5 manipulator with simple velocity control to perform experiments, and achieve high success rates. Moreover, we open-source a dataset containing more than 1,500 flight trajectories, including the position and the posture of 6 typical objects.

\end{itemize}


\section{RELATED WORK}
\textbf{Catching in-flight objects.} Accurately trajectory predicting of the flying object is a significant part in the task of catching in-flight objects. Among the in-flight objects, ball is selected as the experimental object in most tasks since its trajectory is considered to be relatively easy to predict. \cite{chen2017hitting,muller2011quadrocopter,riley2002robot} ignore the effect of air drag and other force applying on the ball and approximate its flight trajectory as a parabola. \cite{bauml2011catching} measures the air drag coefficient in the experimental environment and models the ball accurately in dynamics. \cite{dong2020catch,bauml2010kinematically,frese2001off} use the EKF \cite{barker1995bayesian} to correct ball's flight state. Based on these trajectory prediction methods, the above works have achieved good experimental results catching the ball. But  problems come out when these methods are applied to other flying objects.\par

For the prediction of the trajectory of other flying objects, \cite{jia2019batting,gardnermotion} accomplishes this task by dynamical modeling of the flying objects as well as EKF, but the dynamical modeling requires prior information such as mass, position of COM, moment of inertia, and shape of the objects, which need to be modeled separately for each objects. And it is very difficult to collect these information for uneven objects. \cite{kim2014catching,kim2012estimating,salehian2016dynamical} estimate the dynamics model of the object by offline learning through the traditional machine learning method, which achieve decent results in predicting the translation and rotation of the object. However, the model parameters of the traditional learning method will augment as the sample size increases, and in addition, the models obtained by traditional learning have poor performance when generalizing to unseen objects.\par

\begin{figure*}[t]
\centering
\includegraphics[width=\linewidth]{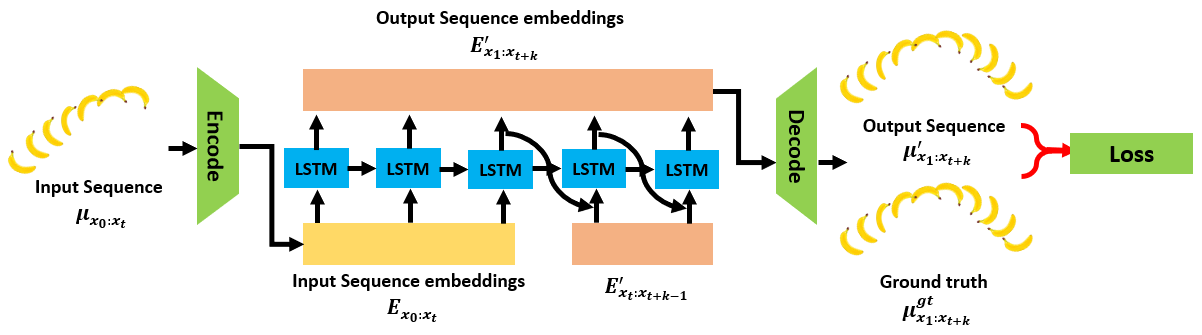}
\caption{NAE, an LSTM based acceleration estimator that fusing information from multiple frames. During training, NAE inputs a object state sequence $\mu_{x_{0}: x_{t-1}}$ from time $0$ to $t$ and outputs prediction $\mu_{x_{1}: x_{t+k}}$ from time $1$ to $t-1+k$. Three loss functions are defined to train the NAE model, respectively for learning single-step prediction, multi-step prediction and embedding reconstruction. During using, NAE inputs a object state sequence $\mu_{x_{0}: x_{t}}$ from time $0$ to $t-1$ to estimate the object state $\mu_{x_{t}}$ at time $t$. }
\label{fig:NAE}
\end{figure*}

\textbf{Differentiable estimator.} Conventional Bayes filters have been widely applied for state estimation in robotic applications \cite{thrun2002probabilistic}. On the other side of front-end, the latest deep learning technique brings more precise measurement with sensors \cite{chen2020survey}. In this context, the combination of the state estimation and deep learning is becoming a research focus in recent years.\par

To achieve this, Haarnoja et al. \cite{haarnoja2016backprop} proposed a backprop Kalman filter system, in which the state estimator and recurrent network are trained together with backpropagation. The authors demonstrated the effectiveness with synthetic tracking task and visual odometry in the real world. As for the range sensors, the differentiable Kalman filter can also be integrated to the end-to-end system for vehicle pose tracking \cite{yin2020rall}. Despite the Kalman filter above, some research works focused on building end-to-end particle filter \cite{jonschkowski2018differentiable,karkus2018particle}. However, since the resampling step is non-differetiable, most of particle filter based methods only used the network as measurement model \cite{yin2018locnet}, and the whole system is not trained in end-to-end manner. In \cite{karkus2018particle}, a soft-sampling policy is introduced to address the issue, thus making the differentiable particle filter feasible for state estimation.

\section{METHODS}

The flight of uneven object is a dynamic system constructed of position, velocity, and acceleration. In some works, all other external forces are not considered. The acceleration is fixed to only gravity, and the order of the system can be reduced, leading to the prediction error. For a better physical model, the acceleration is regarded time-varying, but it lacks real-time measurement, thus is modeled for a specific object in prior, failing to be generalized to new object. Following this idea, our model uses NAE to online measure the acceleration, and update the state, achieving accuracy and generalization at the same time.

In order to deal with the system's non-linearity caused by air drag, air lift and Magnus force,  etc.\cite{gardnermotion}, Neural Acceleration Estimator(NAE), as shown in Fig. \ref{fig:NAE}, explicitly estimates time-varying acceleration by a Long Short-Term Memory network. Taking both prediction result given by linear propagation model and the measurement made by NAE into consideration, we supervise the uncertainty of measurement step by end to end training NAE with a Differentiable Kalman Filter. Fig. \ref{fig:NAE-DF} shows the integral diagram of NAE-DF.  We introduce NAE in Section \ref{NAE} and give more detail about NAE-DF model in Section \ref{NAE-DF}, as shown in Fig. \ref{fig:DKF} \par

\subsection{Neural Acceleration Estimators} 
\label{NAE}
Uneven objects are affected by various kinds of forces, which leads to a non-linear dynamic system. This non-linearity changes acceleration all the time during object's flight. In-flight acceleration is hard to be measured by sensors but can be estimated by observing a small fragment of previous deflected trajectory, which is accomplished by the Neural Acceleration Estimators. 

As a non-linear model, the LSTM network is qualified for fusing information from multiple frames and learning long-term dependence. As shown in Fig. \ref{fig:NAE}, we define the object state  $\mu_{x_{t}}\in\mathbb{R}^{9} $ at frame $t$ as a nine-dimensional vector composed of the current position, velocity and acceleration. The NAE model inputs a object state sequence $\mu_{x_{0}: x_{t}}$ from frame $0$ to $t-1$ to estimate the object state $\mu_{x_{t}}$ at next frame $t$. We first map the input state sequence to a high-dimensional space $E_{x_{t}} \in \mathbb{R}^{128}$ using an Encoder consist of fully connected layers, and use $tanh$ as the activation function. After processing, the embedded output $ \boldsymbol{E}_{x_{t+1}}^{\prime}$ is remapped back to $\boldsymbol{\mu}_{x_{t+1}}^{\prime}$ in the original space through a Decoder as the final result of NAE. The core of NAE is an LSTM-based multi-step prediction model. We use three loss functions for training.\par

\textbf{One-step Teacher forcing loss:} One-step Teacher forcing loss helps NAE learn single-step prediction. Given the sequence of object state $ \mu_{x_{0}: x_{t}} $ from frame $0$ to $t-1$, NAE outputs the sequence $ \mu_{x_{1}: x_{t+1}}^{\prime} $ from frame $1$ to $t$. One-step Teacher forcing minimizes the MSE loss between the prediction sequence $ \mu_{x_{1}: x_{t+1}}^{\prime} $ and the ground truth sequence $\mu_{x_{1}: x_{t+1}}^{g t}$ ,\par

\begin{equation}
\mathcal{L}_{1}=\frac{1}{t} \sum_{i=0}^{t-1}\left\|\mu_{x_{i+1}}^{g t}- \mu_{x_{i+1}}^{\prime}\right\|^{2}
\end{equation}

\textbf{Multi-step Free running loss:} Multi-step Free running helps NAE learn multi-step prediction. Given the sequence of object state $ \mu_{x_{0}: x_{t}} $ from frame $0$ to $t-1$, NAE outputs the object state sequence $ \mu_{x_{1}: x_{t+1+k}}^{\prime} $ from frame $1$ to $t+k$ after $k$ prediction steps. For the frame with a given state, the input of the LSTM unit is the mapping $E_{x_{t}}$ of the known state $\mu_{x_{t}}$ at the corresponding frame, and for the frame without given state, the input is the output $\boldsymbol{E}_{x_{t-1}}^{\prime}$ of the LSTM unit at the previous frame $t-1$. Multi-step Free running also uses MSE loss to minimize the error between the prediction sequence $ \mu_{x_{t+1}: x_{t+1+k}}^{\prime} $ and the ground truth sequence $ \mu_{x_{t+1}: x_{t+1+k}}^{g t} $,\par

\begin{equation}
\begin{split}
\mathcal{L}_{2}=\frac{1}{k} \sum_{i=t}^{t-1+k}\left\|\mu_{x_{i+1}}^{g t}- \mu_{x_{i+1}}^{\prime}\right\|^{2}
\end{split}
\end{equation}

\textbf{Reconstruction loss:} Reconstruction step learns the Encoder and Decoder composed of a fully connected layers. The Encoder maps the input sequence $\mu_{x_{0}: x_{t}}$ from $0$ to $t-1$ to the high-dimensional space, then the Decoder maps it back to the original space, minimizing the reconstruction loss,\par

\begin{equation}
\mathcal{L}_{3}=\frac{1}{t} \sum_{i=0}^{t-1}\left\|\mu_{x_{i}}^{g t}-D\left(E\left(\mu_{x_{i}}\right)\right)\right\|^{2}
\end{equation}

\subsection{Neural Acceleration Estimator with Differentiable Filter}
\label{NAE-DF}
Uncertainty helps us to make an optimal fusion of prediction and measurement by estimating a joint probability distribution. Since uncertainty of both prediction and measurement are lack of supervision, we embed NAE into a Differentiable Filter, which takes the acceleration estimated by NAE as the measurement as shown in Fig. \ref{fig:NAE-DF}. In the learning stage, we train the NAE-DF in an end-to-end manner and a maximum likelihood formulation.\par

\begin{figure}[t]
\centering
\includegraphics[width=8cm]{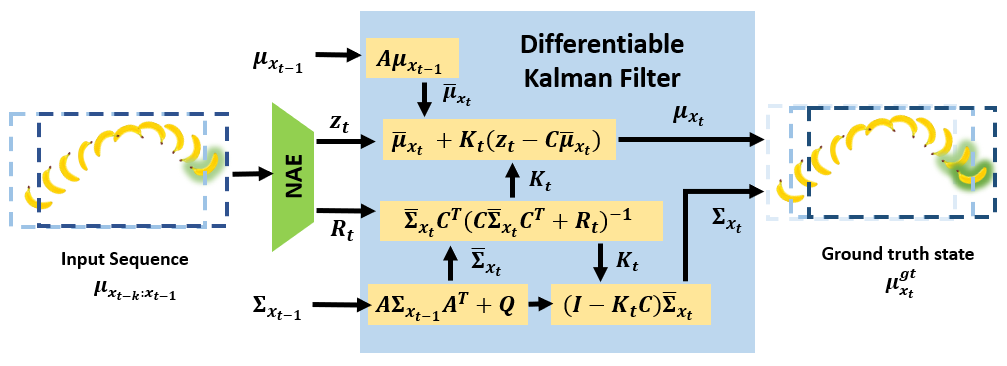}
\caption{The internal details of the Differentiable Filter. The model consists of a prediction step with linear dynamical model and a measurement step with non-linear NAE. }
\label{fig:DKF}
\end{figure}

Kalman Filter models statistical noise and other inaccuracies and is a general technique to fuse the sensor data in sequential\cite{yin2020rall}. In prediction step, known dynamical model will propagate the current state to the next state, during which the uncertainty increases. Then the measurement step makes estimation more confident by multiplying two Gaussian distributions. Fig. \ref{fig:DKF} gives the internal details of the DF module. As we put all complex non-linearity caused by air drag, air lift, Magnus force and gravity, etc. into the NAE's measurement, we can use a simple linear propagation model derived from gravity as the prediction model. Therefore the prediction step is formulated as follows: 
\begin{equation}
\bar{\mu}_{x_{t}}=A \mu_{x_{t-1}}
\end{equation}
\begin{equation}
\bar{\Sigma}_{x_{t}}=A \Sigma_{x_{t-1}} A^{T}+Q 
\end{equation}
where the matrix A is $$ A=\left[\begin{array}{ccccccccc}
1 & 0 & 0 & \Delta t & 0 & 0 & \frac{1}{2} \Delta t^{2} & 0 & 0 \\
0 & 1 & 0 & 0 & \Delta t & 0 & 0 & \frac{1}{2} \Delta t^{2} & 0 \\
0 & 0 & 1 & 0 & 0 & \Delta t & 0 & 0 & \frac{1}{2} \Delta t^{2} \\
0 & 0 & 0 & 1 & 0 & 0 & \Delta t & 0 & 0 \\
0 & 0 & 0 & 0 & 1 & 0 & 0 & \Delta t & 0 \\
0 & 0 & 0 & 0 & 0 & 1 & 0 & 0 & \Delta t \\
0 & 0 & 0 & 0 & 0 & 0 & 1 & 0 & 0 \\
0 & 0 & 0 & 0 & 0 & 0 & 0 & 1 & 0 \\
0 & 0 & 0 & 0 & 0 & 0 & 0 & 0 & 1 \\
\end{array}\right] $$
and $Q$ is a pre-defined noise covariance matrix.\par

For the measurement step, we use NAE as the observation model whose input is a recent sample of object in-flight trajectory and output is an observation of the current state of the object. The covariance matrix $R_{t}$ of the current state is also supervised by end-to-end training. Measurement step is formulated as follows:
\begin{equation}
K_{t}=\bar{\Sigma}_{x_{t}} C^{T}\left(C \bar{\Sigma}_{x_{t}} C^{T}+R_{t}\right)^{-1}
\end{equation}
\begin{equation}
\mu_{x_{t}}=\bar{\mu}_{x_{t}}+K_{t}\left(z_{t}-C \bar{\mu}_{x_{t}}\right)
\end{equation}
\begin{equation}
\Sigma_{x_{t}}=\left(I-K_{t} C\right) \bar{\Sigma}_{x_{t}}
\end{equation}
where $C$ is the observation matrix. Since we only take the acceleration term of the NAE's estimated state as the observation, the matrix C is defined as: $$ C=\left[\begin{array}{ccccccccc}
0 & 0 & 0 & 0 & 0 & 0 & 1 & 0 & 0 \\
0 & 0 & 0 & 0 & 0 & 0 & 0 & 1 & 0 \\
0 & 0 & 0 & 0 & 0 & 0 & 0 & 0 & 1 \\
\end{array}\right]$$\par
From a probability point of view, Differentiable Filter maximizes the data likelihood $\mathcal{N}\left(\mu_{x_{t}},\Sigma_{x_{t}}\right)$, so we can train NAE-DF by maximizing the posterior probability, as shown in Fig. \ref{fig:NAE-DF}:
\begin{equation}
\text { maximize } \mathcal{N}\left(\mu_{x_{t}}^{g t} ; \mu_{x_{t}}, \Sigma_{x_{t}}\right) 
\end{equation}
We apply negative log-likehood to formulate the loss function:
\begin{equation}
\mathcal{L}_{4}=\frac{1}{k} \sum_{t-k}^{t}\left(\mu_{x_{t}}^{g t}-\mu_{x_{t}}\right)^{T} \Sigma_{x_{t}}^{-1}\left(\mu_{x_{t}}^{g t}-\mu_{x_{t}}\right)+\gamma \operatorname{det}\left(\Sigma_{x_{t}}\right)
\end{equation}
where $\gamma$ is a constant factor for regulation. \par

\section{SYSTEM OVERVIEW}

\begin{figure}[t]
\centering
\includegraphics[scale=0.3]{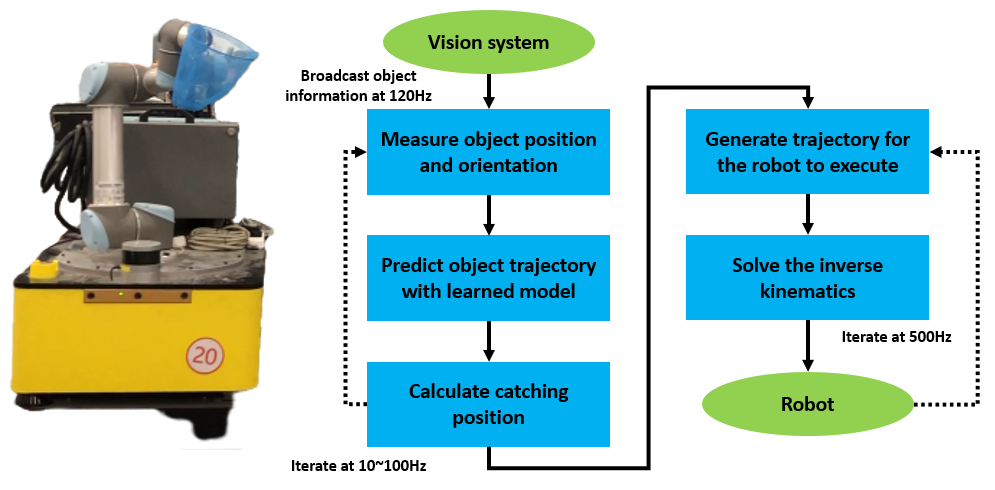}
\caption{The system flow and hardware equipment we use to form our real world catching system. A motion capture system provides vision information including object's position and orientation at 120 Hz. The interception position is calculated by learned model, then the arm following a velocity control law will move to the target position for capture.}
\label{fig:system-overview}
\end{figure}

Fig. \ref{fig:system-overview} illustrates the system flow and hardware equipment of the real world catching system we built to verify our algorithm. First, we collect the flight information of uneven objects through a motion capture system. Next, two independent threads work simultaneously to generate the interception position and control the UR5 respectively. The interception position generation thread predicts the trajectory of the flying object through the model, and generates the appropriate interception position by calculating the intersection of the trajectory and the arm's workspace. According to the generated position, the robot arm control thread calculates the motion trajectory by inverse kinematics, and corrects the position through velocity control during reestimation. The velocity control law is defined as follows:
\begin{equation}
v=\frac{1-e^{-kd}}{1+e^{-kd}}v_{max}
\end{equation}
where $d=\left \| \mathbf{p}_{cur}-\mathbf{p}_{tar}  \right \|^{2}$ is the distance between the arm's current position $\mathbf{p}_{cur}$ and the target position $\mathbf{p}_{tar}$, $k$ is a proportional coefficient used to accelerate the convergence of the UR5, and $v_{max}$ is the maximum velocity of UR5 executor. In our experiment we chose $k=12$ and $v_{max}=1.85m/s$.

\begin{figure}[t]
\centering
\includegraphics[width=7.5cm]{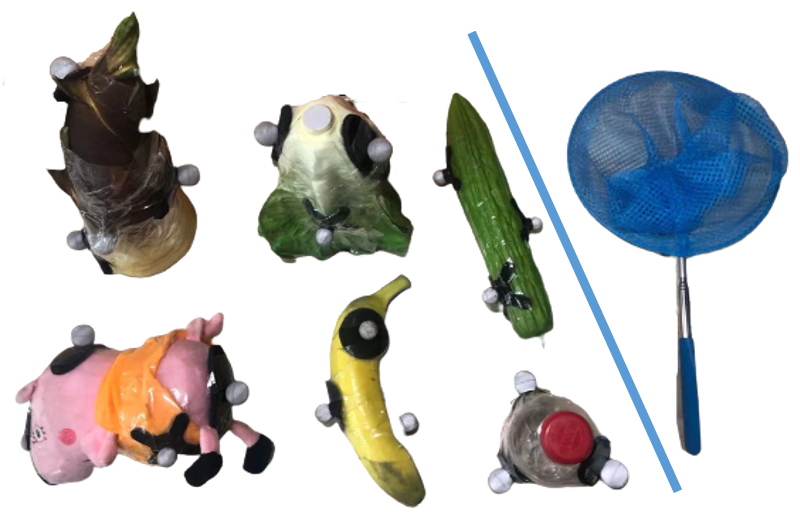}
\caption{The objects in the left are uneven objects we used for data collection. Infrared reflectors are fixed on objects for motion capture. Blue spoon net in the right is used as the basket. The Radius of spoon net is 10 $cm$.}
\label{fig:dataset_object}
\end{figure}

\begin{figure}[!t]
\centering
\includegraphics[width=7.5cm]{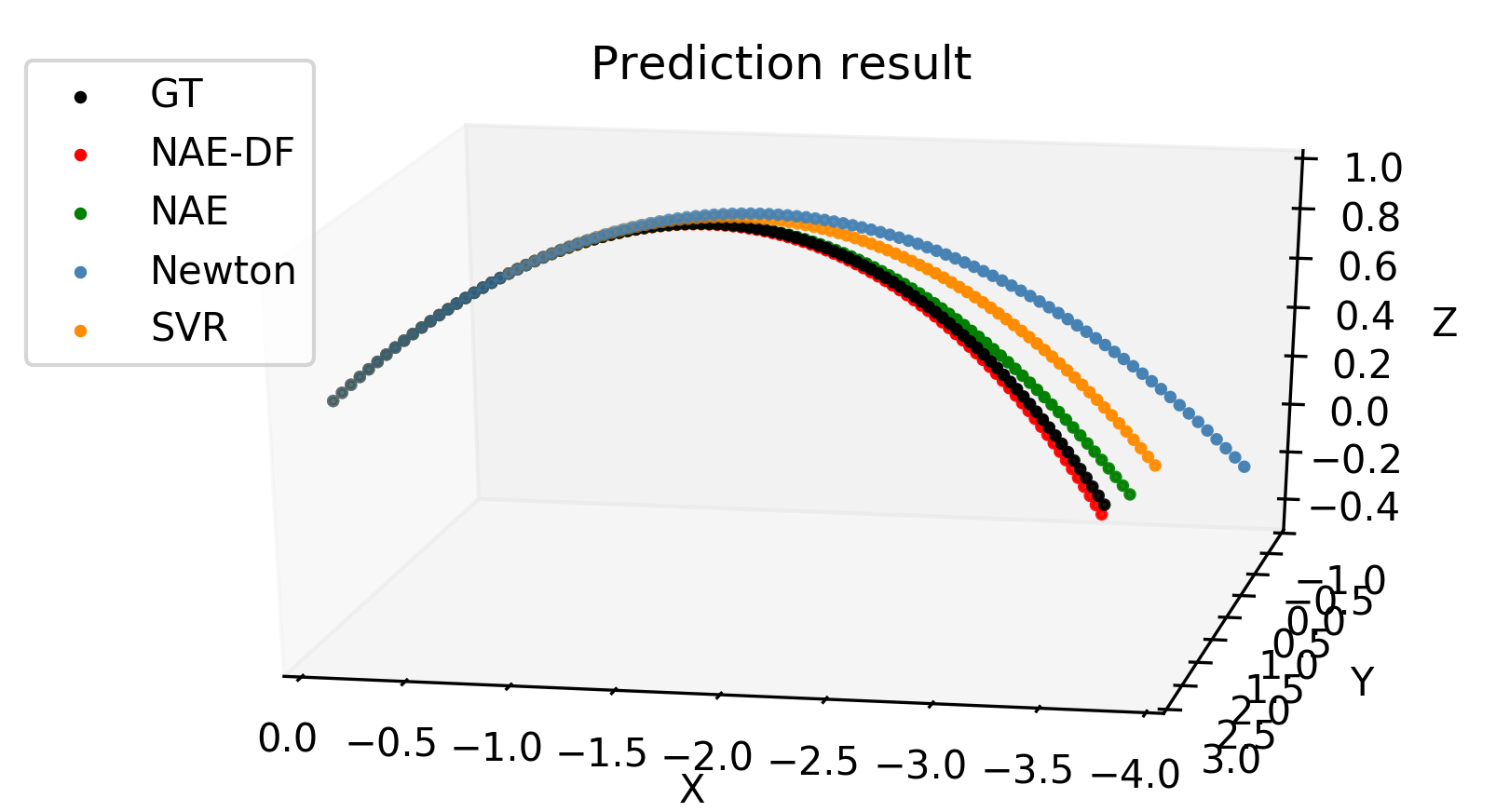}
\caption{Prediction result of the four algorithms for a trajectory in banana testing set. NAE-DF's prediction(red) is the most similar to the ground truth(black) trajectory.}
\label{fig:dataset_prediction}
\end{figure}

\begin{figure}[!t]
\centering
\includegraphics[width=8.5cm]{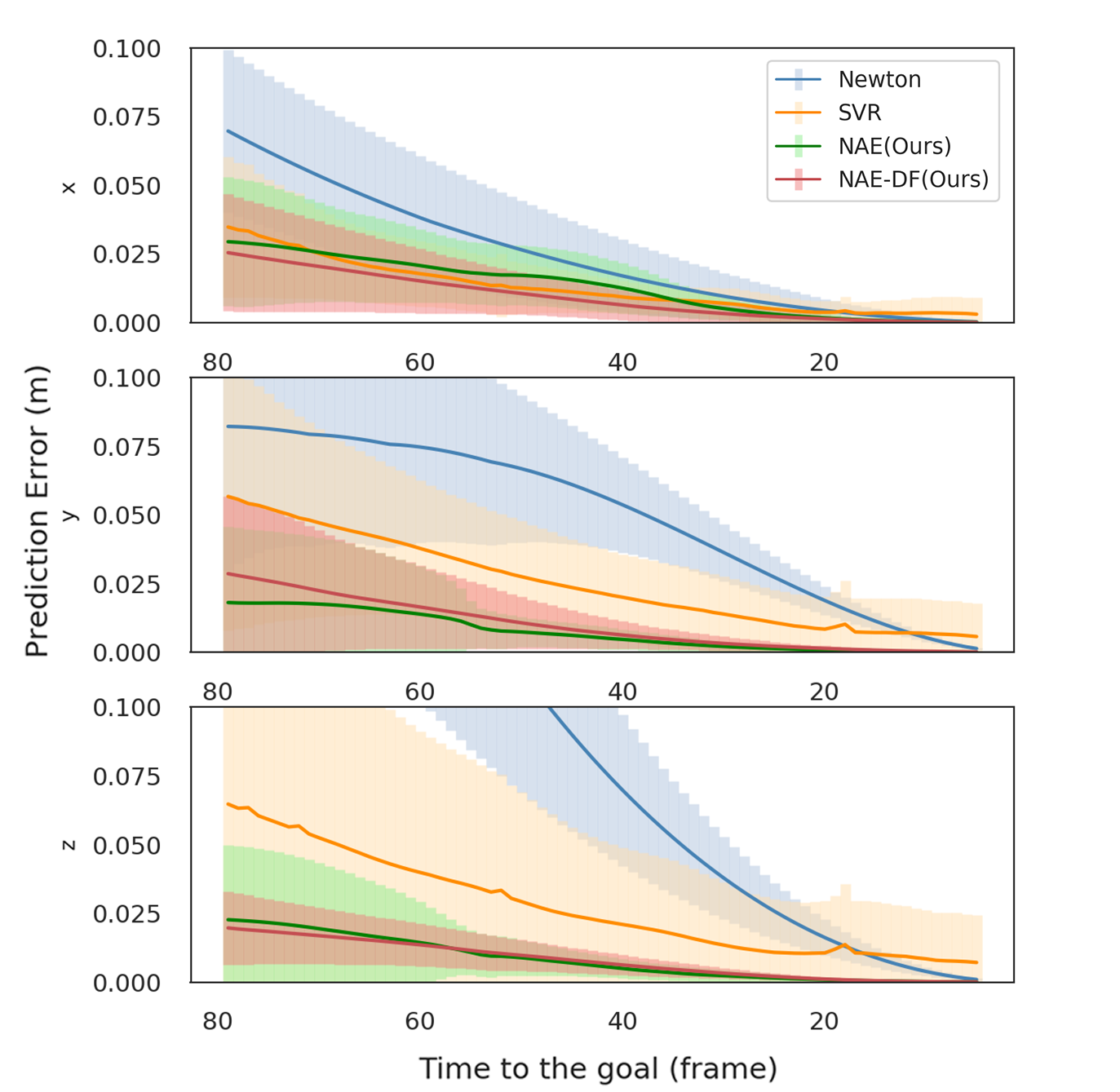}
\caption{The accumulated error decreases when there are fewer frames to predict. Horizontal coordinate represents remained frames to the goal. Vertical coordinate represents the ${L}_{2}$ norm with goal position's coordinates. In the early time, NAE and NAE-DF get smaller accumulated error and standard deviation than the other two methods.}
\label{fig:dataset_xyz_error}
\end{figure}

\begin{figure}[htbp]
\centering
\includegraphics[width=8cm]{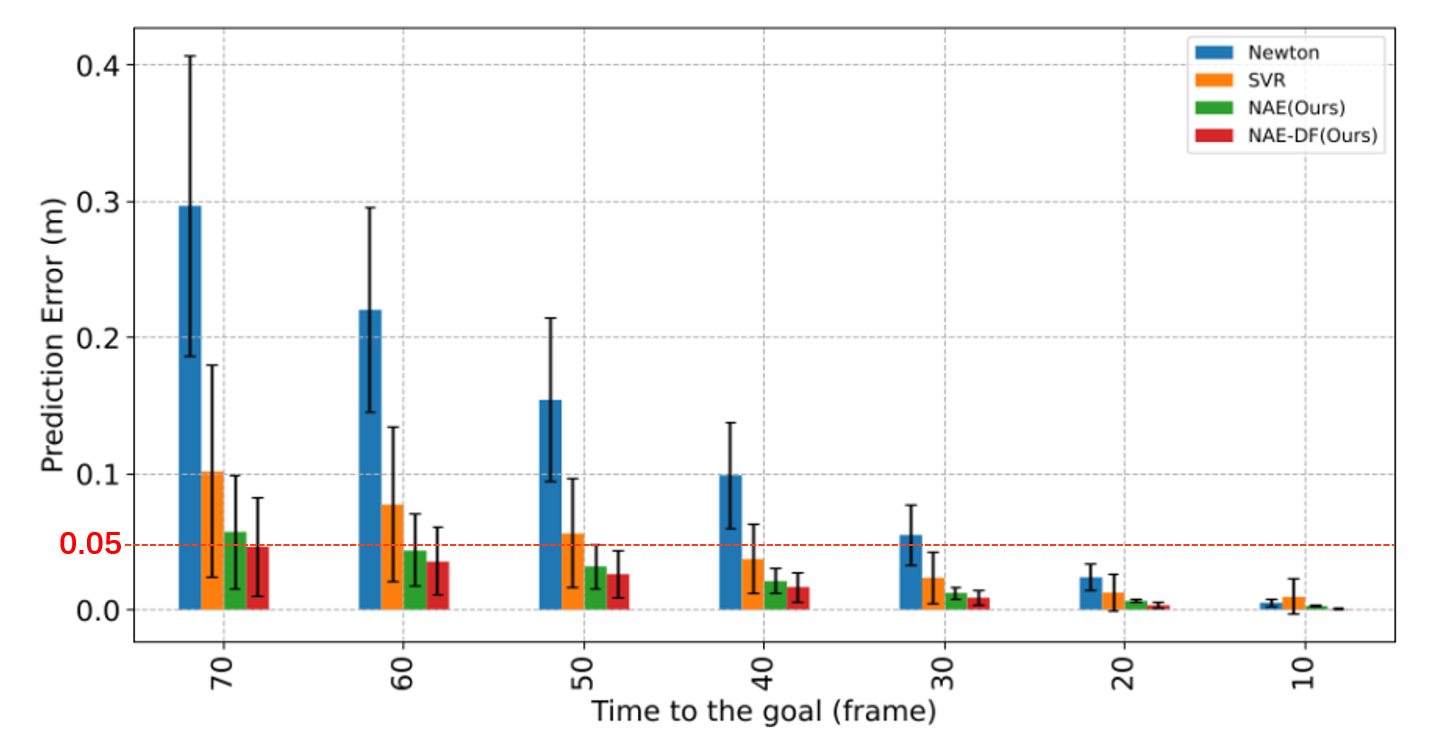}
\caption{The accumulated error decreases when there are fewer frames to predict. Horizontal coordinate represents remained frames to the goal. Vertical coordinate represents the ${L}_{2}$ norm with ground truth of goal position. In the early time, NAE and NAE-DF get smaller accumulated error and standard deviation than the other two methods. NAE-DF's error stays under 0.05m.}
\label{fig:dataset_Banana}
\end{figure}

\begin{table*}[htbp]
\renewcommand\arraystretch{1.5}
\centering
\caption{Leading Time Criterion defined by how early achieve the desired precision as 1$cm$ in Cartesian space. Larger value means higher prediction accuracy. NAE-DF gets the highest value, followed by NAE.}
\label{table1}
\begin{tabular}{c|c|c|c|c|c|c}
\hline
\hline
 &
  Bottle(half) &
  Bamboo &
  Banana &
  Green &
  Gourd &
  Paige \\ \hline
Newton &
  0.15$\pm$0.03 &
  0.11$\pm$0.02 &
  0.10$\pm$0.02 &
  0.08$\pm$0.01 &
  0.10$\pm$0.02 &
  0.09$\pm$0.01 \\ \hline
SVR &
  0.12$\pm$0.08 &
  0.14$\pm$0.15 &
  0.16$\pm$0.09 &
  0.04$\pm$0.08 &
  0.09$\pm$0.11 &
  0.14$\pm$0.11 \\ \hline
NAE (Ours) &
  0.21$\pm$0.08 &
  0.23$\pm$0.05 &
  0.25$\pm$0.08 &
  0.17$\pm$0.06 &
  0.25$\pm$0.06 &
  0.26$\pm$0.07 \\ \hline
\textbf{NAE-DF(Ours)} &
  \textbf{0.25$\pm$0.09} &
  \textbf{0.30$\pm$0.08} &
  \textbf{0.30$\pm$0.09} &
  \textbf{0.20$\pm$0.04} &
  \textbf{0.26$\pm$0.07} &
  \textbf{0.28$\pm$0.10} \\ 
  \hline
  \hline
\end{tabular}
\end{table*}

\section{EXPERIMENTAL RESULT}

In this section, we carried out a series of experiments to evaluate our method. The goals of the experiments are: 
\begin{itemize}
\item to demonstrate our method is capable of predicting more accurate trajectories of in-flight uneven objects than traditional learning methods.
\item to indicate that our method is effective and efficient for the task of catching in-flight uneven object with velocity of 5-6m/s.
\item to test the generalization to unseen objects of  our method.
\end{itemize}

We compare our method with two methods: 1) Newton: a method which regards the trajectory of the in-flight uneven objects as a parabola. 2) Support Vector Regression (SVR) \cite{kim2014catching,kim2012estimating}: a method using bundle of machine learning methods to estimate non-linear dynamics learning from some samples of trajectories.

\subsection{Dataset Experiment}
\label{VA}
We collect over 1,500 trajectories of 6 typical objects, 90$\%$ of which for training set and 10$\%$ for testing set. See Fig. \ref{fig:dataset_object} for more details. All vegetable models are made of PU(ploy urethane) materials, including a banana, a bamboo, a green and a bitter gourd. Besides we also have a toy Paige, and a bottle with water. The in-flight dataset are generated with data augmentation methods such as translation and rotation around $Z$ axis so models can adapt to various inputs. We train the SVR, NAE, and NAE-DF models on the training set, and compare their prediction precision on the testing set. In dataset evaluation, the goal of the trajectory is set to the last frame's position.\par 
\textbf{Prediction Result}. Fig. \ref{fig:dataset_prediction} shows the prediction result of the four algorithms on a trajectory in banana's testing set. The diagram shows that the error of direct propagation using gravity acceleration(Newton's method\cite{hong1997experiments,riley2002robot}) is very large, followed by the SVR and NAE. NAE-DF's prediction is the most similar to the ground truth trajectory. Fig. \ref{fig:dataset_xyz_error} and Fig. \ref{fig:dataset_Banana} show the accumulated error decreases when there are fewer frames to predict. In the early time, NAE and NAE-DF get smaller accumulated error and standard deviation than the other two methods. This suggests that our time-varying acceleration estimation is beneficial to improve the trajectory prediction accuracy. Besides, NAE-DF can achieve a better performance than NAE because the end-to-end training with Differentiable Filter can provide stronger constraints.

\textbf{Leading Time Evaluation}. Kim et al.\cite{kim2012estimating} define an evaluation criterion for different algorithms that is how early achieve the desired precision as 1$cm$ in Cartesian space, called Leading Time. This is also the desiderate to compensate lag of execution time for a robot. The larger the value is, the better the model performs. Table. \ref{table1} indicates that for a banana, NAE-DF can obtain an estimation results less than 1$cm$ error with 0.30$s$ ahead, and the value for NAE, SVR, and Newton' method are 0.25$s$, 0.16$s$, and 0.10$s$ respectively. This is because the flight of the uneven object is a highly non-linear system. In constrast, NAE and NAE-DF are capable of modeling this non-linearity thus performs more efficiently in this experiment. \par

\begin{figure*}[t]
	\centering
	\subfigure[SVR Generalization]{
		\includegraphics[width=5.7cm]{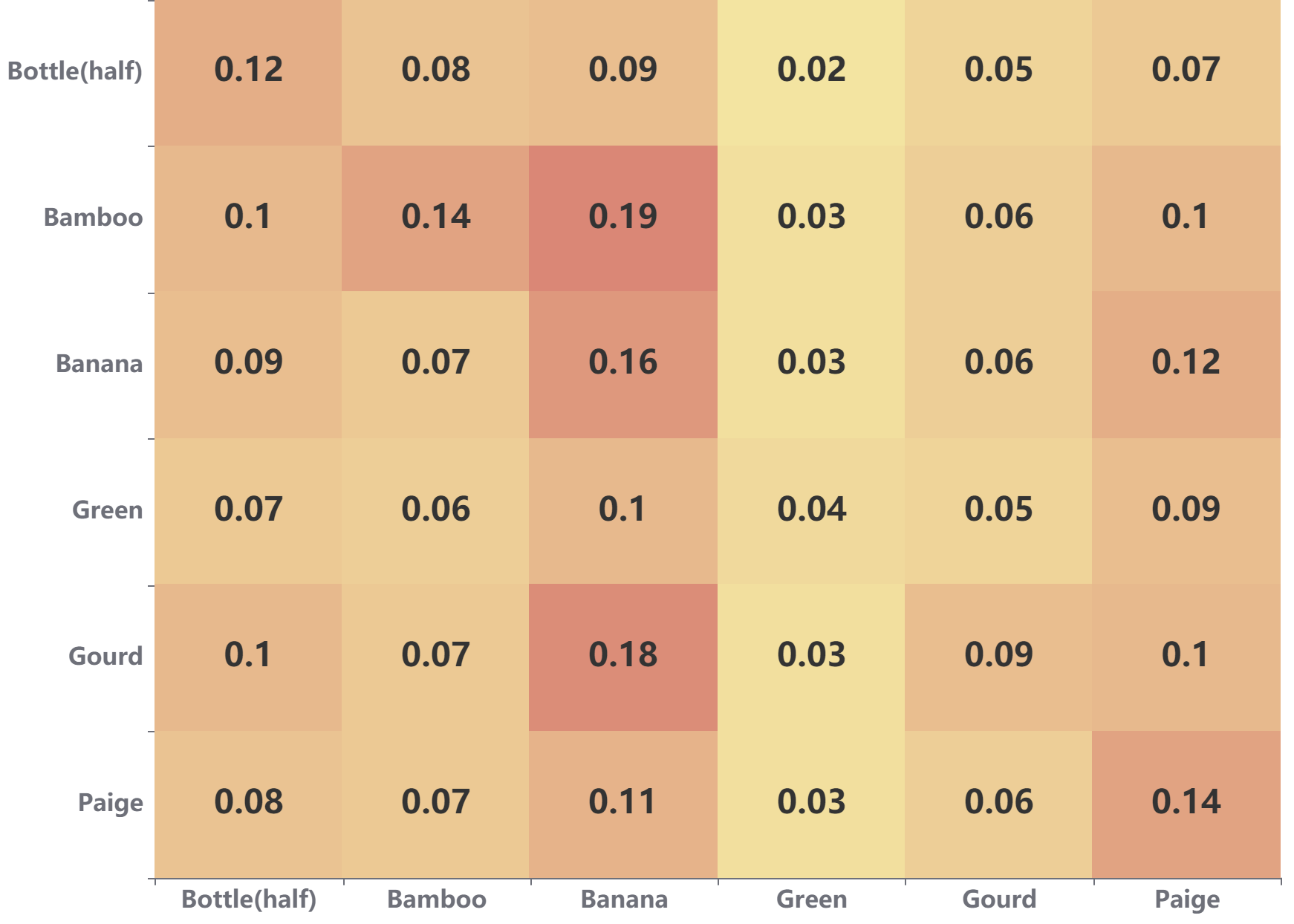}}
	\subfigure[NAE Generalization]{
		\includegraphics[width=5.7cm]{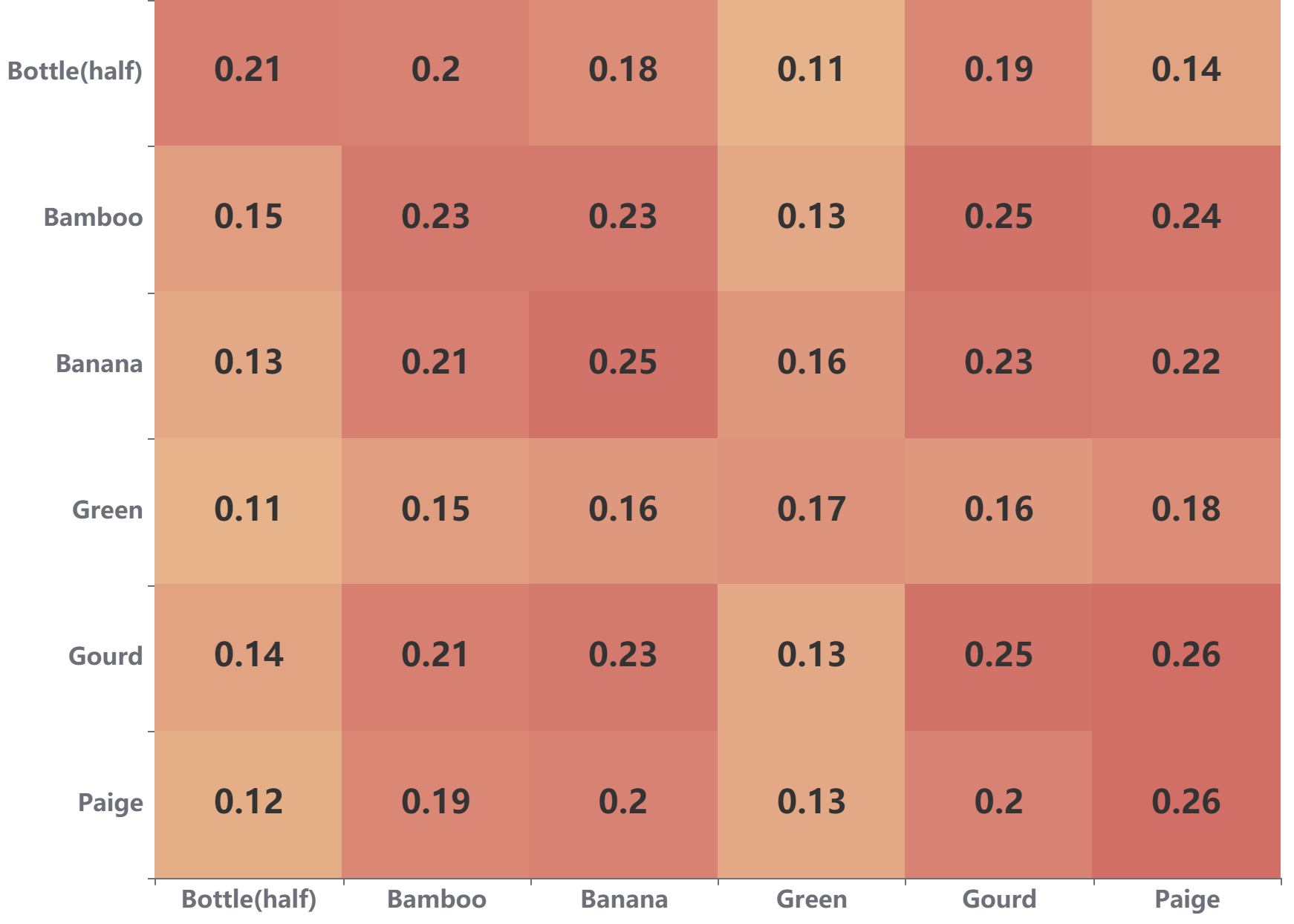}}
	\subfigure[NAE-DF Generalization]{
		\includegraphics[width=5.7cm]{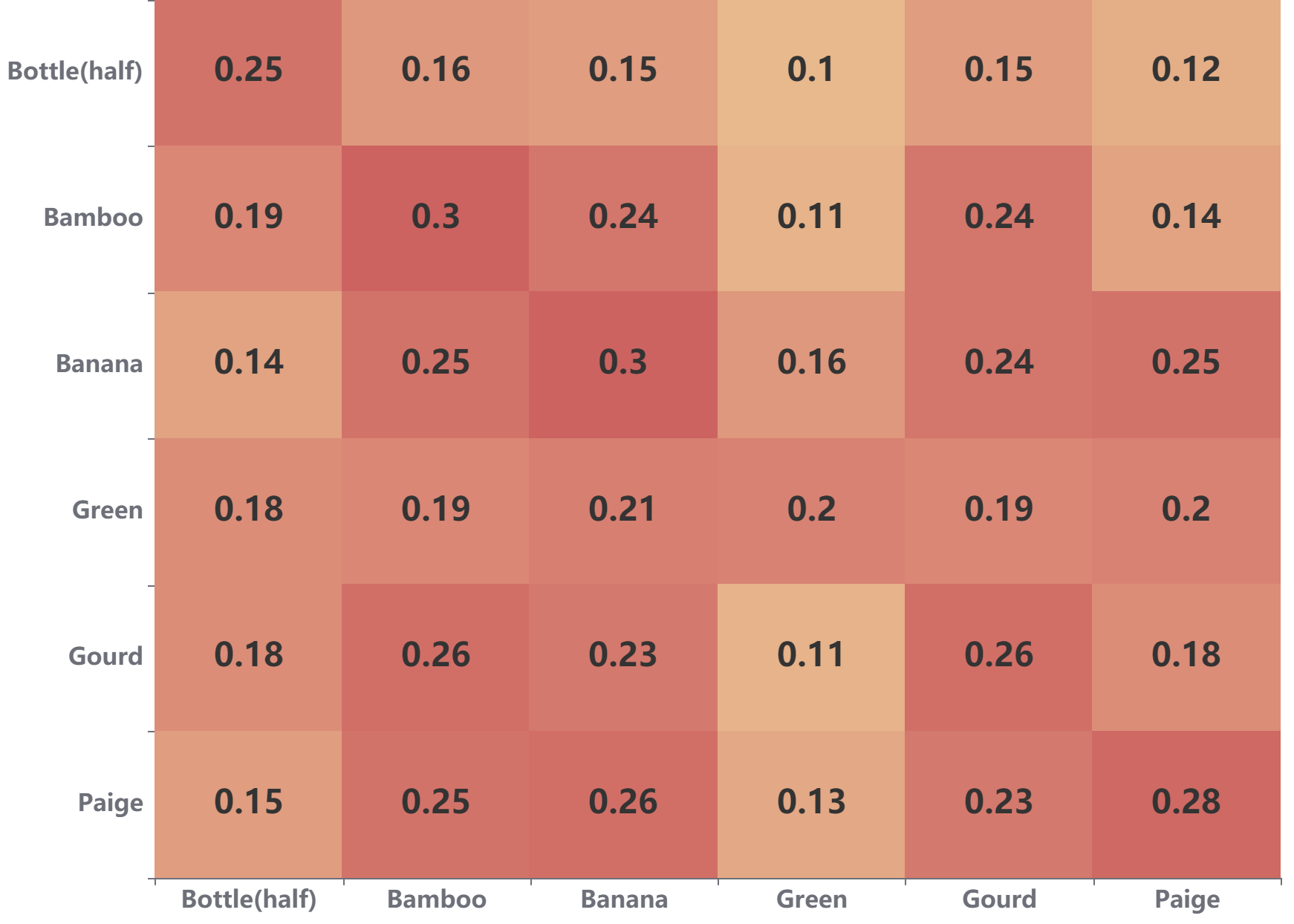}}
	\caption{Generalization performance indicated by Leading Time. $x$ coordinate represents the dataset we test. $y$ coordinate represents the dataset we used to train the model.}
	\label{fig:dataset_gen_error}
\end{figure*}

\textbf{Leading Time Evaluation for Generalization}. In order to verify the generalization performance, we use the Leading Time criterion mentioned above for comparison. As shown in Fig. \ref{fig:dataset_gen_error}, the proposed NAE and NAE-DF are superior to Newton's method and SVR in generalization performance for unseen objects. The generalization performance of NAE-DF on objects other than the bottle is better than NAE. This is due to the inductive bias injected to the learning architecture by NAE and NAE-DF.\par

\begin{figure*}[htbp]
\centering
\includegraphics[width=\linewidth]{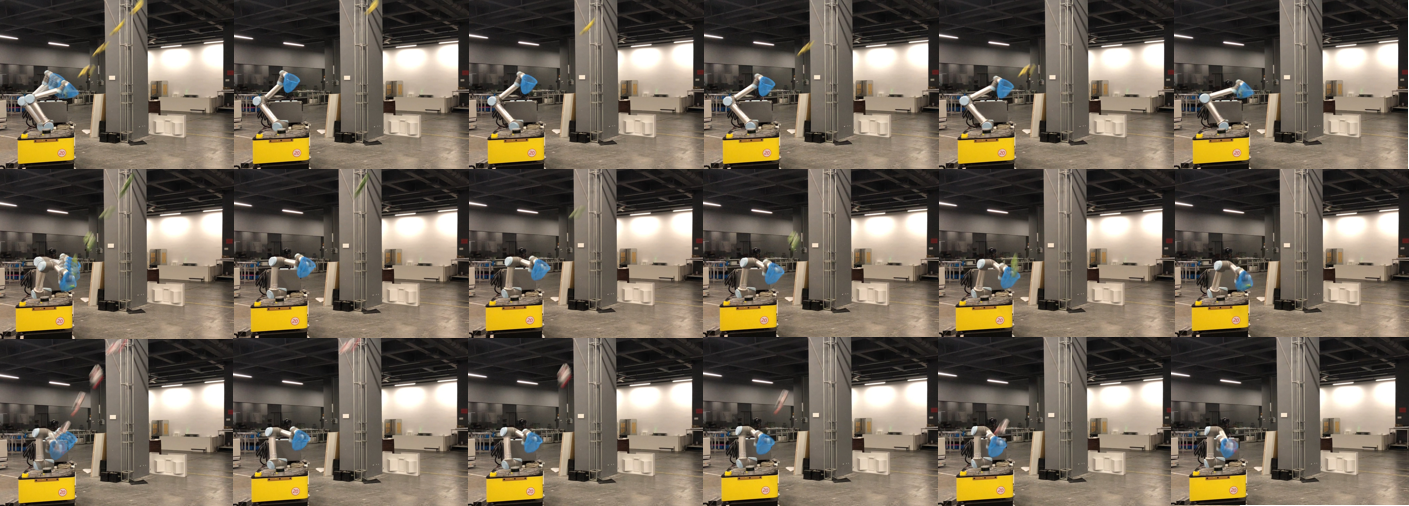}
\caption{Success cases for capture a PU banana, a PU bitter gourd and a bottle with water. The used model is NAE-DF trained on banana's data.}
\label{fig:real_world_seq}
\end{figure*}

\subsection{Real World Experiment}
\label{VB}
Fig. \ref{fig:real_world_seq} gives an illustration of our real world experiments. We try to capture a PU banana, a PU bitter gourd and a bottle with water by the real world system. Except those throws that don't intersect with the arm’s workspace, the NAE-DF model trained on banana's data has a success rate of $83.3\%$ when catching a flying banana, while $73.3\%$ for NAE, $56.7\%$ for SVR, $40.0\%$ for Newton method, which validates effectiveness of our method for catching in-flight uneven objects in real world experiments. Then we test the generalization performance of the model trained with banana data by catching the unseen bitter gourd and a bottle with water. The success rates of each generalization experiments are $86.7\%$ and $53.3\%$ respectively, indicating that our model has a decent generalization performance.

\subsection{Failure Case Analysis}
\label{VC}
\begin{figure*}[htbp]
\centering
\includegraphics[width=\linewidth]{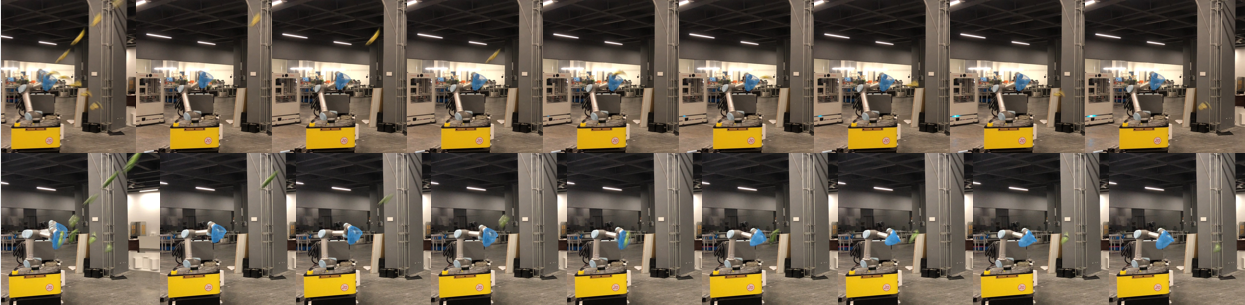}
\caption{Typical failure case. Objects are bounced off by the basket for a sub-optimal interception pose.}
\label{fig:real_world_failure}
\end{figure*}

The typical failure cases in the real world experiment are shown in the Fig. \ref{fig:real_world_failure}. The main reason that leads to a failure is that currently we don't consider the optimal interception pose, but only the interception position. The radius of the basket is relatively small (10$cm$, see Fig. \ref{fig:dataset_object} for more detail) for some objects' poses. In these cases, objects will be bounced off by the basket. In the following work, we will carry out the motion prediction for in-flight pose, which will optimize the interception pose.

\section{CONCLUSIONS}

Tackling the challenging task of motion prediction for in-flight uneven object, we propose a Nerual Acceleration Estimator that measures the time-varying acceleration caused by system's non-linearity. Furthermore, we embed NAE into a 
 Differentiable Filter which is trained in the end-to-end manner to give a supervision to uncertainty in measurement. 
 We verify the effectiveness of the algorithm on the dataset as well as on the real world system. With simple velocity control, it is possible to achieve interception of a variety of objects. Compared with the existing uneven objects' motion prediction methods, our algorithms have great advantages in prediction accuracy and generalization performance. At the same time, we open-source an in-flight object dataset. In the future work, we will predict the flying pose of the objects and optimize the motion planning of the robotic arm for further improvement.

\printbibliography

\end{document}